\documentclass[runningheads]{llncs}
\usepackage{graphicx}
\usepackage{bbm}
\usepackage{amsfonts}

\begin{document}
\title{Leveraging Automated Mixed-Low-Precision Quantization for tiny edge microcontrollers}
\titlerunning{Leveraging Automated Quantization for MCUs}
\author{Manuele Rusci\inst{1,2} \and
Marco Fariselli\inst{2} \and
Alessandro Capotondi\inst{3} \and 
Luca Benini \inst{1,4}}
\authorrunning{M. Rusci, M. Fariselli, A. Capotondi, L. Benini}
\institute{Università di Bologna, Bologna, ITA \and
Greenwaves Technologies, Bologna, ITA \and
Università di Modena e Reggio Emilia, Modena, ITA \and
IIS ETH Zurich, Zurich, CH\\
\email{manuele.rusci@unibo.it}}

\maketitle              % typeset the header of the contribution
\begin{abstract}
The severe on-chip memory limitations are currently preventing the deployment of the most accurate Deep Neural Network (DNN) models on tiny MicroController Units (MCUs), even if leveraging an effective 8-bit quantization scheme.
To tackle this issue, in this paper we present an automated mixed-precision quantization flow based on the HAQ framework but tailored for the memory and computational characteristics of MCU devices. 
Specifically, a Reinforcement Learning agent searches for the best uniform quantization levels, among 2, 4, 8 bits, of individual weight and activation tensors, under the tight constraints on RAM and FLASH embedded memory sizes.
We conduct an experimental analysis on MobileNetV1, MobileNetV2 and MNasNet models for Imagenet classification. 
Concerning the quantization policy search, the RL agent selects quantization policies that maximize the memory utilization.
Given an MCU-class memory bound of 2MB for weight-only quantization, the compressed models produced by the mixed-precision engine result as accurate as the state-of-the-art solutions quantized with a non-uniform function, which is not tailored for CPUs featuring integer-only arithmetic. This denotes the viability of uniform quantization, required for MCU deployments, for deep weights compression.
When also limiting the activation memory budget to 512kB, the best MobileNetV1 model scores up to 68.4\% on Imagenet thanks to the found quantization policy, resulting to be 4\% more accurate than the other 8-bit networks fitting the same memory constraints.

\end{abstract}
\section{Introduction}
%%%%%  scaling to the edge  %%%%%  
Tiny smart devices feature low-end processing units to interpret sensor data and extract meaningful and compressed information. Among the digital processing solutions, Micro-Controller Units (MCUs) are highly desirable because of high flexibility, due to the software programmability, low-cost and ultra-low power consumption, which can be as low few $mW$, hence compatible with the requirements of battery-operated edge sensor systems. However, MCUs feature a tightly bound on-chip memory budget (mostly for cost reasons), i.e. typically not more than a few MB of internal flash storage and 1MB of RAM. Such a \emph{memory bottleneck} stands as the major limitation for bringing the state-of-the-art inference Deep Learning (DL) models on these devices~\cite{banbury2020benchmarking}. For instance, the MobileNetV1~\cite{howard2017mobilenets} model features up to 4.24~M parameters, resulting in a 16~MB of weight storage (32-bit floating point, i.e. FP32, format), which is much higher than the typical size of on-chip memories.  

\begin{figure}[t]
    \centering
    \includegraphics[width=\columnwidth]{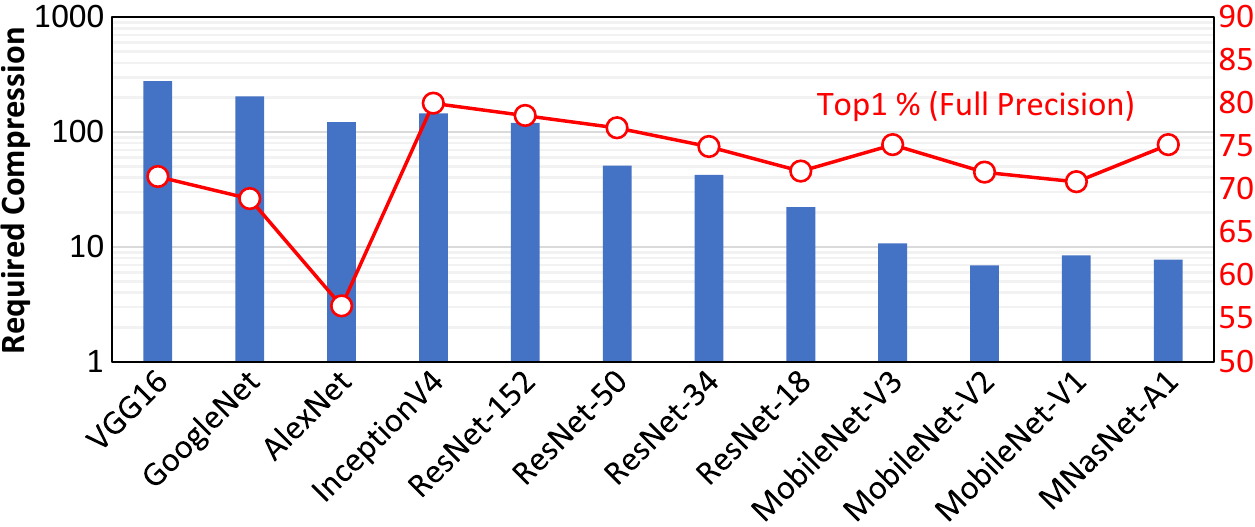}
    \vspace{-.3cm}
    \caption{The histogram (left axis) shows the required model compression of popular state-of-the-art CNNs for fitting in 2MB. The red line (right axis) displays the Top1\% on Imagenet task for the same models~\cite{canziani2016analysis}.}
    \label{fig:motivational}
    \vspace{-.5cm}
\end{figure}

Figure~\ref{fig:motivational} plots the required compression ratio to be applied to several Imagenet classification models~\cite{canziani2016analysis} for fitting into a memory budget of 2~MB. To reach the goal, a $>$100$\times$ compression factor is required for highly accurate models, e.g. InceptionV4 and ResNet-152, while optimized topologies, e.g. Mobilenets, demand a $\sim$10$\times$ compression. Typically, quantization is used to shrink a DL model at the cost of an accuracy penalty with respect to the full-precision counterpart. Recent works demonstrated that 8-bit quantization applies almost losslessly but leads only to a 4$\times$ compression over full-precision models (FP32). Therefore, to meet such requirement, a sub-byte quantization, i.e. using less than 8-bit, must be applied at the cost of a (potential) non-negligible accuracy loss~\cite{choi2018pact,courbariaux2016binarized}. 
%
%%%%%  mixed low precision and CmixNN %%%%%%%%
To reduce the accuracy degradation when applying aggressive quantization, Mixed-Low-Precision quantization techniques have been introduced~\cite{rusci2019memory,haq}. Differently from homogeneous quantization, which relies on a network-wise \textit{Q}-bit compression, a mixed-precision scheme defines an individual bitwidth for every weights and activation tensors of a deep model~\cite{rusci2019memory}, namely the \textit{quantization policy}. Hence, an effective mixed-precision quantization framework aims at finding the best quantization policy that leads to the highest accuracy under given memory and computational constraints. 
While Rusci et. al.~\cite{rusci2019memory} proposed a rule-based, but accuracy-agnostic, mechanism to determine the quantization policy based on the memory characteristics of the target MCU device, Wang et al. proposed HAQ~\cite{haq}, which automates the search phase by means of a Reinforcement Learning (RL) agent to effectively explore the accuracy vs quantization policy space. Unfortunately, the HAQ study relied on a clustering-based, hence non-linear, quantization scheme when optimizing for memory compression, which cannot be effectively mapped on the integer-only arithmetic of low-end micro processors.

%%%%%  Automation for MCUs%%%%%  
In this work, we improved the automated mixed-precision framework~\cite{haq} for targeting the tight computational and memory constraints of a tiny MCU device. 
Besides the on-chip memory limitations, we constrained the quantization policy search to ensure a solution that efficiently maps on the software backends \textit{PULP-NN}~\cite{bruschi2020enabling} or \textit{CMix-NN}~\cite{capotondi2020cmix}, the leading solutions for mixed-low-precision deployment on RISCV and ARM-Cortex M devices, respectively. Specifically: (a) the search is conducted under the memory constraints dictated by both FLASH and RAM memory sizes, which impact respectively, the weights and activation feature maps compression, (b) instead of non-linear quantization, the RL agent enforces an \emph{Uniform} quantization optimized for tiny MCU devices and (c) the tensor bit precision is restricted to the set of \emph{power-of-2} formats (2-, 4-, 8-bit), the formats supported by the target software libraries.

%%%%%  Contributions  %%%%%  
The main contributions of this paper are:
\begin{itemize}
    \item We present an automated mixed-precision quantization flow constrained by the HW/SW characteristics of the target tiny MCU device.
    \item We quantify the impact of Uniform Linear quantization within the HAQ framework given the memory and computational constraints.
    \item We apply the automated mixed-precision on the more efficient models for Imagenet classification, i.e. MobileNetV1~\cite{howard2017mobilenets}, MobileNetV2~\cite{sandler2018mobilenetv2} and MNasNet~\cite{tan2019mnasnet}, under the typical MCU memory budget.
\end{itemize}
When experimenting DL models for Imagenet classification optimized for mobile deployments, such as MobileNetV1, MobileNetV2, and MNasNet, we show that a Uniform weight-only quantization strategy, also combined with a restricted bitwidth selection, results to be lossless if compared to the optimal non-Uniform quantization featured by the HAQ framework.
For both cases, we observed the RL agent converging towards quantization policy solutions that fill the target memory budget.
When introducing the quantization of activations, the highest accuracy has been measured on MobileNetV1 with a Top1 of 68.4\% on ImageNet under a memory constraint of 512kB of RAM and 2MB of FLASH, which is 4\% higher than the best 8-bit uniform quantized network that fits the same memory budget (MNasNet 224\_0.35).

\section{Related Work}
Recent hand-crafted model design for image classification task has been driven from efficiency rather than accuracy metrics only. MobileNetV1~\cite{howard2017mobilenets}, MobileNetV2 \cite{sandler2018mobilenetv2} or ShuffleNet~\cite{zhang2018shufflenet} are relevant examples of deep networks that trade-off classification errors with respect to model size and execution latency.
More recent Neural Architecture Search (NAS) methodologies automate the model design process, employing Reinforcement Learning (RL)~\cite{tan2019mnasnet} or backpropagation, i.e. Diffentiable NAS~\cite{liu2018darts}.
% quantization but not sufficient to fit memory constraints
This work is complementary to this class of studies: besides model compression, low-bitwidth quantization reduces the computational and memory bandwidth requirements concerning the full-precision model, e.g. up to 4$\times$ smaller and faster in case of 8-bit quantization. Efficient models, such as MobileNetV1, have been turned into an 8-bit integer-only representation with an almost negligible accuracy loss through a quantization-aware retraining process~\cite{jacob2018quantization}, which is currently state-of-the-art for resource-constrained devices. However, 8-bit quantization is not sufficient to fit the tiny memory budgets of MCUs. 
On the other side, going below 8 bits on both weights and activation values demonstrated to be challenging~\cite{zhou2016dorefa}. Krishnamoorthi et al.~\cite{krishnamoorthi2018quantizing} employed per-channel quantization rather than per-tensor to recover accuracy degradation when lowering the precision to 4 bits. In the case of extreme single-bit quantization, the accuracy gap concerning full-precision networks is still high, preventing their usage on complex decision problems~\cite{hubara2016binarized,rastegari2016xnor}.
All these works feature homogeneous quantization, which is sub-optimal if targeting the compression for resource-constrained devices.

    % mixed-precision (HAQ,Rusci,Hessian) - connection between the first two
Motivated by the low on-chip memory budgets of low-cost MCUs systems, the SpArSe framework~\cite{fedorov2019sparse} exploits pruning to fit the RAM and ROM memory budgets for, respectively, intermediate activations values and weights parameters. Rusci et al.~\cite{rusci2019memory} addressed the same deployment problem on more complex Imagenet classification models using low-bitwidth mixed-precision quantization. In this approach, the selection of the quantization bitwidith is driven by a heuristic rule-based process aiming at fitting the memory constraints. This process resulted effective by showing a 68\% MobileNetV1 model fitting the 2MB FLASH and 512kB RAM constraints of an MCUs. However, the optimality of this quantization approach is limited by being accuracy-agnostic.
On the other side, HAWQ-V2~\cite{dong2019hawq} individuates the bit precision of weights and activations based on a layer-wise hessian metric, which quantifies the sensitivity of any parameter to low-bitwidth quantization concerning the loss.
More interestingly, the HAQ framework~\cite{haq} exploits an RL actor to explore the mixed-precision quantization space given a memory constraint. The framework reported state-of-the-art accuracy metric for quantized MobileNetV1 and MobileNetV2 models under memory constraints of less than 2MB. However, the framework makes use of optimal non-uniform quantization when optimizing for memory, which makes the solution not suitable for integer-only inference required for MCU's deployment.
Moreover, HAQ exploits arbitrary quantization formats (in the range between 2- to 8-bit), which are not effectively supported by most of the MCU backend library for CNNs.
In this work, we extend HAQ for deployment on MCU, hence introducing uniform quantization for both weights and activation values for mixed-precision quantization policy search under the MCU's memory constraints.

\section{Automated Mixed-Precision Quantization for MCU} \label{sec:uhaq}
Given a pretrained full-precision model, the presented mixed-precision flow produces a tensor-wise quantization policy that 1) leads to the lowest accuracy drop with respect to the full-precision model and 2) matches the computational and memory constraints of the target MCU architecture. 
In this section, after formulating the optimization objectives and briefly describing the state-of-the-art HAQ framework~\cite{haq}, we describe the improvements applied to HAQ to quantize a DL model for deployment into a tiny MCU device.
\subsection{MCU-aware optimization objectives}  \label{sec:problem}
An automated mixed-precision quantization framework for MCUs operates with a double scope. On one side, (i) the memory requirements of the quantized model must fit the on-chip memory resources of the target MCU and, at the same time, (ii) the quantized model must be linearly quantized to low-bitwidth to be efficiently processed on a general-purpose CPU featuring integer-only arithmetic.
\subsubsection{Memory Requirement}
MCU's on-chip memory distinguishes between RAM and ROM memories. In this context, embedded FLASH memories are classified as ROM memories, because writing operations does not typically occur at runtime. 
As a design choice, we store the weight parameters in the ROM memory, while intermediate activations values are stored in RAM~\cite{rusci2019memory,chowdhery2019visual,fedorov2019sparse}. Hence, we derive the following constraints:
\begin{enumerate}
    \item[M1] The memory requirement due to weight parameters, including bias or other stationary values, must fit the system's ROM memory ($\textrm{M}_{ROM}$). 
    \item[M2] The RAM memory must store any intermediate activation tensors at runtime. In addition to the input and output tensors of any network layer, tensors of optional parallel branches (e.g. skip connections) must be also kept in memory ($\textrm{M}_{RAM}$).
\end{enumerate}
\subsubsection{Computational Requirement}
Low-bitwidth fixed-point inference tasks typically operate on 8- or 16-bit data~\cite{lai2018cmsis}. 
%Concerning an ARM Cortex-M CPU, 
The leading software backends \textit{PULP-NN}~\cite{bruschi2020enabling} (RISCV) and \textit{CMix-NN}~\cite{lai2018cmsis} (ARM Cortex M) for mixed-precision inference supports 2- and 4-bit datatypes in addition to the widely used 8-bit compression. 
Operations on sub-bytes datatypes are software-emulated using 8-bit or 16-bit instructions with low overhead. 
%A recent solution for mixed-precision inference, including 2-,4- and 8-bit formats, has been also presented for RISC-V tiny devices~\cite{bruschi2020enabling}.
On the other hand, efficient software implementations for other quantization levels, i.e. 3-, 5-, 6-, 7-bits, have not been presented.
Hence, we restrict the quantization bitwidth selection to \textit{power-of-2} format (2-, 4- and 8-bit) and exploit \textit{Uniform} linear quantization rules to enable the usage of integer-only arithmetic for computation. This latter implies that quantization levels are uniformly spaced across the quantization range~\cite{jacob2018quantization}.
\begin{figure}[t]
    \centering
    \includegraphics[width=0.9\columnwidth]{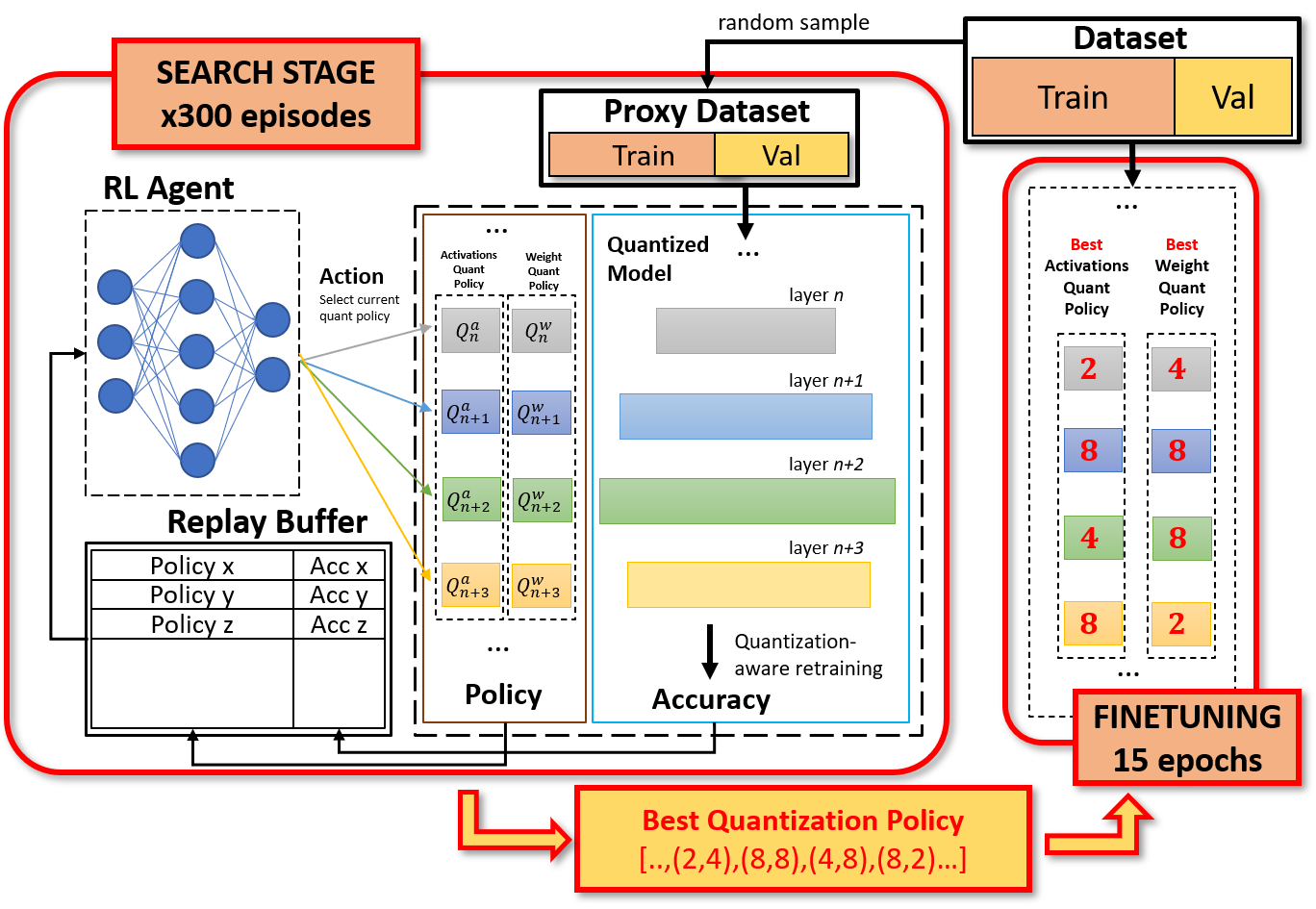}
    \caption{Overview of the search and fine-tuning steps in the HAQ framework~\cite{haq}.}
    \label{fig:HAQ}
    \vspace{-0.05cm}
\end{figure}
\subsection{Automated Precision Tuning}
The framework HAQ~\cite{haq} is the leading automated methodology to optimally select the quantization bitwidth of individual weight and activation tensors given memory and computational constraints.
HAQ works as a two-stage process: (i) a first \textit{Search} stage, driven by a Reinforcement Learning agent to find the best quantization policy followed by a (ii) \textit{Fine-Tuning} stage, where the model parameters are updated by a quantization-aware training process.
Figure~\ref{fig:HAQ} illustrates the main modules of the HAQ engine. 
The \textit{Search} stage occurs over a fixed number of episodes. At any episode, a NN-based RL agent selects a quantization policy for the target DL model and triggers a quantization-aware training process on a proxy dataset for few epochs (1 epoch in our setting). The proxy dataset is a subset of the original dataset, composed by randomly sampled points.
The scored accuracy is returned to the RL agent as the reward function and stored within a replay buffer. Based on these collected data, the RL agent try to learn the relation among the quantization policy and the accuracy over the proxy-dataset. So doing, the RL agent adapts over the episodes and improves its ability to select the best quantization policy, i.e. the one that maximizes the network accuracy while matching the memory or performance requirements.
To guarantee a sufficient action-reward diversity, the RL generates a random policy during an initial batch of episodes, i.e. the warm-up period. After this, the RL agent starts to produce policies based on historical data (the internal model predictions).
After a fixed number of episodes (300 in our setting), the search engine outputs the quantization policy that leads to the highest score on a proxy dataset (the \textit{best quantization policy}).
This latter feeds the \textit{Fine-Tuning} stage, which applies a quantization-aware training over a longer training period (15 epochs in our setting).

\subsection{HAQ for MCU deployments}
When optimizing for memory size, the work \cite{haq} exploits a non-Uniform quantization rule based on a KMeans clustering algorithm. Such a quantization rule maps the real weight parameters in a finite subset of $\mathbb{R}$, composed by $2^Q$ centroids, where $Q$ is the number of bits. Hence, if storing the centroid values into a per-tensor lookup table, any parameter value can be coded as a $Q$-bit integer index. 
This quantization approach results not compatible with the low-bitwidth integer-only arithmetic featured by our target SW backends: as described in Section~\ref{sec:problem}, either activation values, and weights must be uniformly quantized to exploit instruction-level parallelism. Additionally, the resource-constrained devices can lack floating-point support, making a full-precision clustering-based compression not deployable~\cite{flamand2018gap}. 
Therefore, to bridge this gap, we extended the HAQ framework by introducing the following major improvements.

\textbf{Uniform Weight and Activation Quantization}. Weights parameters are uniformly quantized using a linear Per-Channel mapping~\cite{rusci2019memory}. Activations values are also quantized tensor-wise by means of a linear rule. To this aim, fake-quantization layers are inserted at the input of any convolutional layer~\cite{jacob2018quantization} to learn the dynamic range through backpropagation~\cite{choi2018pact} during training. Only \textit{power-of-2} bit precision levels (2-, 4-, 8-bit) are considered by the engine.

\textbf{Weight and Activation Bitwidth Selection}.
During the \textit{Search stage}, the RL agent can operate according to two strategies: (a) a \textit{Concurrent search}, where the activation and weight bitwidths are selected concurrently at every episode, or (b) a \textit{Independent search}, where the weight quantization-policy is firstly found by keeping the activation at full-precision and then a second search finds the best quantization policy for the activation tensors, under this weight quantization setting. 
%The \textit{Concurrent search} is more challenging than the \textit{Independent search} due to the higher number of bitwidth parameters to search.
In addition, for more complex network structures like MobileNetV2 and MNasNet, we found the compression of the residual activation layers as the main culprits of the model accuracy degradation. Due to their importance in the system and the limited memory footprint ($<$15\% of RAM occupancy for the biggest network configurations, i.e. 224$\times$224 input dimension and width multiplier of 1), we set them up with fixed 8-bit quantization.

\textbf{Memory Constraints}.
Given the targeted memory model, the RL agent can independently check the memory constraints \textit{M1} and \textit{M2} against, respectively, the weight and activation bitwidth that are selected at any episode. 
%The framework assumes the existence of at least one policy that satisfies both \textit{M1} and \textit{M2}. 
Concerning the weight parameters, if the total memory requirement of the selected policy exceeds the available $\textrm{M}_{ROM}$ memory, the bit precision of the largest weights tensor is reduced (i.e. from 8 to 4-bit or from 4 to 2-bit). Such a process is repeated until the \textit{M1} constraint is satisfied. 
On the other side, any computational node satisfies the \textit{M2} constraint if the occupation of its input and output tensors fits the RAM memory budget. 
Note that parallel skip connections present in residual layers, increase the memory requirements as they have to be preserved in memory during the computation of the parallel branch of the graph.

\section{Experimental Results}
The conducted experiments evaluate the impact of the proposed automated quantization methodology on the precision of state-of-the-art topologies for image classification when compressed to fit the typical MCU memory budget constraints. 
To identify and quantify these effects we divide the experiments in two steps. First, we focus on the evaluation of the uniform quantization process on the weight parameters with a restricted bit choice (\textit{power-of-2} formats 2-, 4-, 8-bit) instead of an optimal non-uniform rule. Second, we evaluate the accuracy achieved on a state-of-the-art edge MCU device when we apply the automated mixed-precision methodology on both activation and weight values with a uniform linear rule.

\subsection{Experimental Setup}
The mixed-precision framework is evaluated over efficient state-of-the-art models for Imaganet classification, such as MobileNetV1~\cite{howard2017mobilenets}, MobileNetV2~\cite{sandler2018mobilenetv2} and MNasNet~\cite{tan2019mnasnet}. 
In our experiments, the quantization policy search is constrained by the memory targets $\textrm{M}_{ROM}=2MB$ and $\textrm{M}_{RAM}=512kB$, corresponding to the memory footprint of a typical high-performance MCU SoC, such as an STM32H7.
The proxy-dataset used for the \textit{Search stage} is composed of 20k ImageNet random samples for the training and 10k samples for validation.
The training procedure of both the quantization policy search and the quantization-aware fine-tuning makes use of an ADAM optimizer with a learning rate of $1\times10^{-4}$. 
During the search phase, the RL agent selects a random policy for the initial 60 episodes, also denoted as the warm-up period. At any episode, the target model is initialized with pre-trained full-precision parameters and trained by means of a quantization-aware process on the proxy Imagenet.
\begin{table}[t]
    \centering
    \caption{Accuracy after the \textit{Fine-Tuning} of Weight-Only Quantized Models under a memory constraint of 2MB}
    \resizebox{0.9\columnwidth}{!}{\begin{tabular}{l|c|c|c|c|c}
         {\bfseries Model} & 
         {\bfseries \#} & 
         {\bfseries No Quant} & 
         {\bfseries Kmeans} & 
         {\bfseries Kmeans} &
         {\bfseries Uniform PC} 
         \\
         
         & 
         {\bfseries params}&
         {\bfseries FP32} & 
         {\bfseries INT-\{2,..8\}} & 
         {\bfseries INT-\{2,4,8\}} & 
         {\bfseries INT-\{2,4,8\}}  
         \\

         \hline
         \hline
         
         MobileNetV1  & 
         4.24M  & 
         70.6\% &   % FP32
         69.0\% &   % Kmeans-all
         69.6\% &   % Kmeans-2,4,8
         70.1\%    % PC-2,4,8
         \\
         
         MobileNetV2 & 
         3.47M &
         72.0\% &  % FP32
         71.4\% &  % Kmeans-all
         71.3\% &  % Kmeans-2,4,8
         70.6\%   % PC-2,4,8
         \\
         
         MNasNet-A1  & 
         3.9M &
         75.2\% &    % FP32
         69.8\% &   % Kmeans-all
         72.2\% &   % Kmeans-2,4,8
         67.1\%    % PC-2,4,8
         \\
    \end{tabular}}
    \label{tab:weight_only}
    \vspace{-0.3cm}
\end{table}

\begin{figure}[t]
    \centering
    \includegraphics[width=0.8\columnwidth]{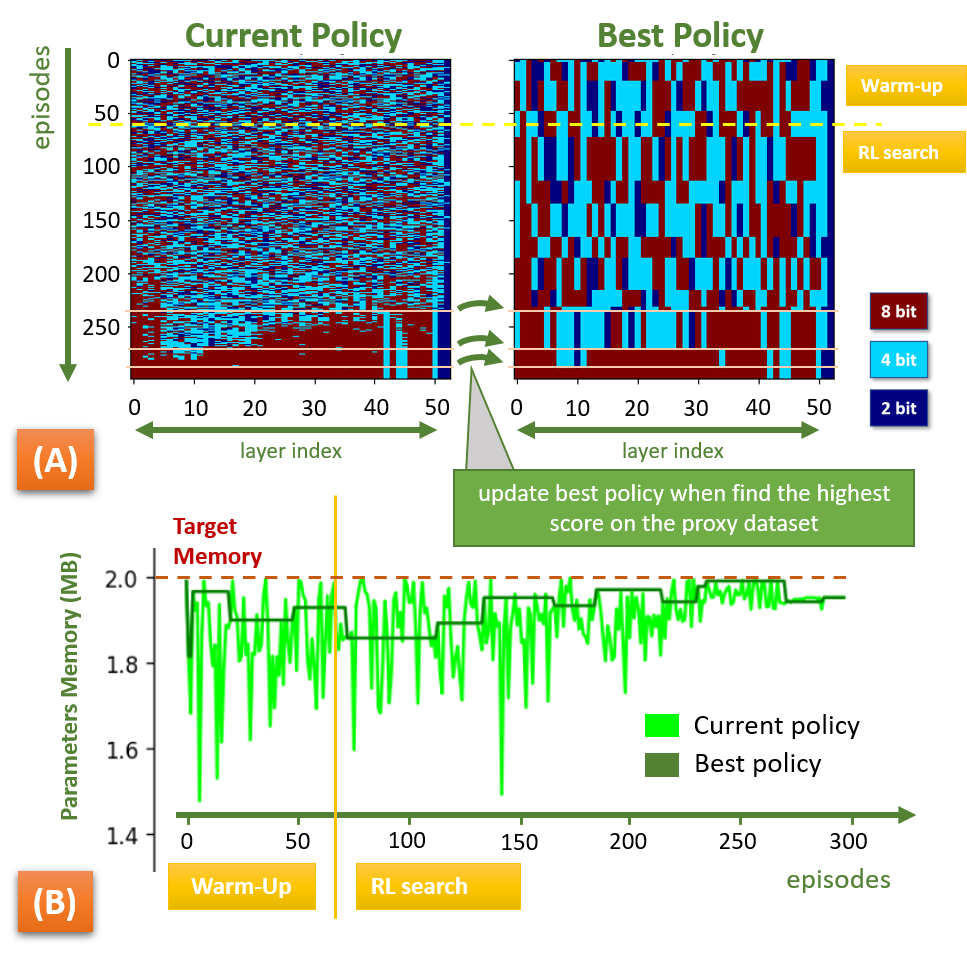}
    \caption{RL-based search of the quantization policy for MobileNetV2 given a memory constraint of 2MB. The plots shows (A) the bit-precision and (B) the memory occupation of the current and best policies selected by the RL agent at every episode.}
    \label{fig:mobv2-wonly-rlsearch}
\end{figure}

\subsection{Automated search for Weight-Only Quantization Policies}
In this section, we analyze the impact on the classification accuracy when running the mixed-precision search of the weight parameters bitwidth with an uniform quantization rule, under the memory constraint $\textrm{M}_{ROM}=2MB$ (activations are kept in full-precision).
Specifically, we compare our uniform weight quantization settings featuring a limited choice of bit precision levels (2,4,8 bits) with respect to a non-uniform optimal compression scheme (based on Kmeans clustering), either with constrained and unconstrained bitwidth selection. This last setting corresponds to the baseline HAQ~\cite{haq} framework. 

% here the example
%The evaluation is conducted over a two-step process: 1) the search phase returns the best quantization policy based on the provided constraints, and afterwards 2) the accuracy of the model quantized with the found mixed-precision policy is assessed on the validation set after a fine-tuning process.
Figure~\ref{fig:mobv2-wonly-rlsearch}A graphically shows the quantization space explored by the agent when searching the best policy for MobileNetV2 with an Uniform quantization scheme. 
The left plot depicts the evolution over the episodes (y-axis) of the weight quantization bit precision at any layer (x-axis), i.e. the current policy, as selected by the RL agent. 
In the plot, bit-precision levels are represented by different colours. 
The plot on the right shows the evolution of the best policies over the episodes: the best policy value is updated when the current policy scores the best Top1 accuracy on the proxy-dataset.
Figure \ref{fig:mobv2-wonly-rlsearch}B plots the memory occupation over the episodes of the target model when featuring the current policy or the best quantization policy. We find the memory footprint of best and current policies to converge towards the memory target (2MB).
This behaviour is observed for all the considered models, demonstrating either that (i) higher Top1 scores are achieved with higher available memory resources and (ii) the RL agent learns to explore more densely the solutions close to the target.
Overall, the memory footprint of the best policies found during the search phase fits the 97th percentile of the memory objective. 

Table~\ref{tab:weight_only} reports the Top1 accuracy scores on Imagenet after the \textit{Fine-Tuning} stage under multiple quantization settings.
When restricting the number of bits to \textit{power-of-2} format, the search space is smaller ($3^l$ vs $7^l$ possible combinations with a \textit{l}-layers network) and the RL problem gets simpler, therefore the agent can find better quantization policies with higher accuracy scores (MobileNetV1 +0.6\% and MNasNet +2.3\%) within the same number of episodes.
When combining the bitwidth restrictions to a Uniform quantization rule, MobileNetV1 shows the same performance level with respect to an unrestricted Kmeans quantized model.
On MobileNetV2 and MNasNet, the accuracy drop is, respectively, limited to -0.8\% and -2.1\%.

\subsection{Automated search for Weight and Activation Quantization Policies}
We evaluate either the \textit{Concurrent} or the \textit{Independent} search approaches to determine the optimal bit precision of both weights and activations under the memory constraints $\textrm{M}_{ROM}=2MB$ and $\textrm{M}_{RAM}=512kB$.
We experimentally observed that both the \textit{Concurrent search} and the \textit{Independent search} outputs the same quantization policies for both MobileNetV2 and MNasNet. 
However, to make the \textit{Concurrent search} converge on the proxy-dataset we found effective to double the number of warm-up and total episodes, from 300 to 600 episodes, with a warm-up period lasting 120 episodes.
Also for the activation policy search, we observed the RL agent  selects activation bitwidths that maximize the available memory.
Table~\ref{tab:acc-act&w} reports the Top1 accuracies of the fully-quantized models after \textit{Fine-tuning} using the mixed-precision flow together with the achieved compression factors, either for weights and activations, against the full precision model. A uniform quantization scheme is used either for the search and fine-tuning stages. 
We run experiments by varying the input resolution $\rho$=\{224, 192, 160\} and width multiplier $\alpha$=\{0.75, 1.0\}  of the target models (indicated by $\rho\_\alpha$ in the following). 
The MobileNetV1 $224\_1.0$ shows the best accuracy, reaching up to 68.4\% on Imagenet. 
On the contrary, MobileNetV2 and MNasNet $224\_1.0$ demand a high compression ratio on the initial layers due to the memory overhead to store the residual values. 
This leads to a non-negligible accuracy drop, respectively of 17\% and 13\% with respect to the full-precision models.  
For model configurations with smaller input dimensions, the memory constraint on the activation can be met with less aggressive quantization strategies. This results in a smaller accuracy drop with respect to the full-precision model.
For this reason, MobileNetV2 $160\_1.0$ reaches an overall accuracy of 67.5\% after fine-tuning, only -1.3\% lower with respect to the full-precision model.
Overall, our solutions show up to +4\% Top1 on Imagenet with respect to an 8-bit MNasNet-A1 $224\_0.35$ model that fits the same memory constraints, even assuming a lossless compression if compared to the full-precision model.

\begin{table}[t]
    \centering
    \caption{Top1 accuracy scores of mixed-precision quantized models with $\textrm{M}_{ROM}=2MB$ and $\textrm{M}_{RAM}=512kB$}
    \resizebox{\columnwidth}{!}{\begin{tabular}{l|c|c|c|c|c}
         & \textbf{\#Params} & \textbf{FP32} & \textbf{Mixed Prec.} & \textbf{Mixed Prec.}& \textbf{Mixed Prec.}\\
         
          \textbf{Model}  & {(M)} & {Top1} & {Top1} & {Weight} & Activ. {(average) }\\
           & {} & {Accuracy} & {Accuracy} & {Compression} & {Compression}\\

         \hline
         \hline

         MobileNetV1 (224\_1.0) & 4.2 & 70.6\% & \textbf{68.4\%} &  9x & 5.1x\\
         MobileNetV1 (224\_0.75) & 2.6 & 68.4\%  & 68.0\% &  5.4x   &  4.6x\\
         \hline
         MobileNetV2 (224\_1.0) & 3.4 & 72.04\% & 55.1\% & 7.6x & 5.1x \\
         MobileNetV2 (192\_1.0) & 3.4 & 70.7\%  & 58.0\% &  7.4x & 5.2x\\
         MobileNetV2 (160\_1.0) & 3.4 & 68.8\% & 67.5\% &  7.4x & 4.4x \\
%         MobileNetV2 (224\_0.75) & 2.6 & 69.8\% & 56.1\% &  5.8x & 7.8x\\
         \hline
         MNasNet-A1 (224\_1.0) & 3.9 & \textbf{75.2\%} & 62.6\% & 9.4x & 4.7x\\
         \hline
         \hline
         MNasNet-A1 (224\_0\_35) & 1.7 & 64.1\% & 64.1\%* & 4x (8-bit) & 4x (8-bit) \\ 
    \end{tabular}}
    * optimistic score: we assume a lossless 8-bit compression w.r.t. FP32 

    \label{tab:acc-act&w}
    \vspace{-0.3cm}
\end{table}

\section{Conclusion}
This work presented a framework that automates the mixed-precision quantization policy search for MCU-constrained targets. The solutions found by our engine tends towards the maximization of the memory objective. 
If applied to state-of-the-art DL models, we demonstrated that uniform mixed-precision weight quantization with restricted \textit{power-of-2} bitwidth does not degrade the accuracy with respect to a non-uniform compression and can be deployed on low-end MCU. Moreover, we reported a fully-quantized mixed-precision network, scoring up to 68.4\% with the best mixed-precision quantization policy found, resulting to be up to +4\% than the best 8-bit network fitting the same memory constraints. 
%However, the non-negligle accuracy degradation shown when applying uniform quantization to both activation and weights indicates a lack on training algorithms for 
\section*{Acknowledgments} 
Authors thank the Italian Supercomputing Center CINECA for the access to their HPC facilities.

\bibliographystyle{splncs04}
\bibliography{main}

\begin{thebibliography}{10}
\providecommand{\url}[1]{\texttt{#1}}
\providecommand{\urlprefix}{URL }
\providecommand{\doi}[1]{https://doi.org/#1}

\bibitem{banbury2020benchmarking}
Banbury, C.R., Reddi, V.J., Lam, M., Fu, W., Fazel, A., Holleman, J., Huang,
  X., Hurtado, R., Kanter, D., Lokhmotov, A., et~al.: Benchmarking tinyml
  systems: Challenges and direction. arXiv preprint arXiv:2003.04821  (2020)

\bibitem{bruschi2020enabling}
Bruschi, N., Garofalo, A., Conti, F., Tagliavini, G., Rossi, D.: Enabling
  mixed-precision quantized neural networks in extreme-edge devices. In:
  Proceedings of the 17th ACM International Conference on Computing Frontiers.
  pp. 217--220 (2020)

\bibitem{canziani2016analysis}
Canziani, A., Paszke, A., Culurciello, E.: An analysis of deep neural network
  models for practical applications. arXiv preprint arXiv:1605.07678  (2016)

\bibitem{capotondi2020cmix}
Capotondi, A., Rusci, M., Fariselli, M., Benini, L.: Cmix-nn: Mixed
  low-precision cnn library for memory-constrained edge devices. IEEE
  Transactions on Circuits and Systems II: Express Briefs  \textbf{67}(5),
  871--875 (2020)

\bibitem{choi2018pact}
Choi, J., Wang, Z., Venkataramani, S., Chuang, P.I.J., Srinivasan, V.,
  Gopalakrishnan, K.: Pact: Parameterized clipping activation for quantized
  neural networks. arXiv preprint arXiv:1805.06085  (2018)

\bibitem{chowdhery2019visual}
Chowdhery, A., Warden, P., Shlens, J., Howard, A., Rhodes, R.: Visual wake
  words dataset. arXiv preprint arXiv:1906.05721  (2019)

\bibitem{courbariaux2016binarized}
Courbariaux, M., Hubara, I., Soudry, D., El-Yaniv, R., Bengio, Y.: Binarized
  neural networks: Training deep neural networks with weights and activations
  constrained to+ 1 or-1. arXiv preprint arXiv:1602.02830  (2016)

\bibitem{dong2019hawq}
Dong, Z., Yao, Z., Cai, Y., Arfeen, D., Gholami, A., Mahoney, M.W., Keutzer,
  K.: Hawq-v2: Hessian aware trace-weighted quantization of neural networks.
  arXiv preprint arXiv:1911.03852  (2019)

\bibitem{fedorov2019sparse}
Fedorov, I., Adams, R.P., Mattina, M., Whatmough, P.N.: Sparse: Sparse
  architecture search for cnns on resource-constrained microcontrollers. arXiv
  preprint arXiv:1905.12107  (2019)

\bibitem{flamand2018gap}
Flamand, E., Rossi, D., Conti, F., Loi, I., Pullini, A., Rotenberg, F., Benini,
  L.: Gap-8: A risc-v soc for ai at the edge of the iot. In: 2018 IEEE 29th
  International Conference on Application-specific Systems, Architectures and
  Processors (ASAP). pp.~1--4. IEEE (2018)

\bibitem{howard2017mobilenets}
Howard, A.G., Zhu, M., Chen, B., Kalenichenko, D., Wang, W., Weyand, T.,
  Andreetto, M., Adam, H.: Mobilenets: Efficient convolutional neural networks
  for mobile vision applications. arXiv preprint arXiv:1704.04861  (2017)

\bibitem{hubara2016binarized}
Hubara, I., Courbariaux, M., Soudry, D., El-Yaniv, R., Bengio, Y.: Binarized
  neural networks. In: Advances in neural information processing systems. pp.
  4107--4115 (2016)

\bibitem{jacob2018quantization}
Jacob, B., Kligys, S., Chen, B., Zhu, M., Tang, M., Howard, A., Adam, H.,
  Kalenichenko, D.: Quantization and training of neural networks for efficient
  integer-arithmetic-only inference. In: Proceedings of the IEEE Conference on
  Computer Vision and Pattern Recognition. pp. 2704--2713 (2018)

\bibitem{krishnamoorthi2018quantizing}
Krishnamoorthi, R.: Quantizing deep convolutional networks for efficient
  inference: A whitepaper. arXiv preprint arXiv:1806.08342  (2018)

\bibitem{lai2018cmsis}
Lai, L., Suda, N., Chandra, V.: Cmsis-nn: Efficient neural network kernels for
  arm cortex-m cpus. arXiv preprint arXiv:1801.06601  (2018)

\bibitem{liu2018darts}
Liu, H., Simonyan, K., Yang, Y.: Darts: Differentiable architecture search.
  arXiv preprint arXiv:1806.09055  (2018)

\bibitem{rastegari2016xnor}
Rastegari, M., Ordonez, V., Redmon, J., Farhadi, A.: Xnor-net: Imagenet
  classification using binary convolutional neural networks. In: European
  Conference on Computer Vision. pp. 525--542. Springer (2016)

\bibitem{rusci2019memory}
Rusci, M., Capotondi, A., Benini, L.: Memory-driven mixed low precision
  quantization for enabling deep network inference on microcontrollers. arXiv
  preprint arXiv:1905.13082  (2019)

\bibitem{sandler2018mobilenetv2}
Sandler, M., Howard, A., Zhu, M., Zhmoginov, A., Chen, L.C.: Mobilenetv2:
  Inverted residuals and linear bottlenecks. In: Proceedings of the IEEE
  Conference on Computer Vision and Pattern Recognition. pp. 4510--4520 (2018)

\bibitem{tan2019mnasnet}
Tan, M., Chen, B., Pang, R., Vasudevan, V., Sandler, M., Howard, A., Le, Q.V.:
  Mnasnet: Platform-aware neural architecture search for mobile. In:
  Proceedings of the IEEE Conference on Computer Vision and Pattern
  Recognition. pp. 2820--2828 (2019)

\bibitem{haq}
Wang, K., Liu, Z., Lin, Y., Lin, J., Han, S.: Haq: Hardware-aware automated
  quantization with mixed precision. In: IEEE Conference on Computer Vision and
  Pattern Recognition (CVPR) (2019)

\bibitem{zhang2018shufflenet}
Zhang, X., Zhou, X., Lin, M., Sun, J.: Shufflenet: An extremely efficient
  convolutional neural network for mobile devices. In: Proceedings of the IEEE
  Conference on Computer Vision and Pattern Recognition. pp. 6848--6856 (2018)

\bibitem{zhou2016dorefa}
Zhou, S., Wu, Y., Ni, Z., Zhou, X., Wen, H., Zou, Y.: Dorefa-net: Training low
  bitwidth convolutional neural networks with low bitwidth gradients. arXiv
  preprint arXiv:1606.06160  (2016)

\end{thebibliography}

\end{document}